\documentclass[letterpaper, 10 pt, conference]{ieeeconf}  %

\IEEEoverridecommandlockouts                              %

\usepackage{soul}

\renewcommand{\baselinestretch}{0.95}
\let\labelindent\relax

\usepackage{algorithm}
\usepackage[noend]{algpseudocode}
\usepackage{float}
\usepackage{bbm}
\usepackage{graphicx}
\usepackage{subcaption}
\usepackage{hyperref}       %
\usepackage{url}            %
\usepackage{booktabs}       %
\usepackage{amsfonts}       %
\usepackage{amsmath,amssymb}
\usepackage{nicefrac}       %
\usepackage{enumitem}
\usepackage{wrapfig}
\usepackage[table]{xcolor}

\newcommand{\Skip}[1]{}

\newcommand{\RNum}[1]{\uppercase\expandafter{\romannumeral #1\relax}}

\title{\LARGE \bf
TWIST: Teacher-Student World Model Distillation for Efficient Sim-to-Real Transfer
\author{Jun Yamada, %
Marc Rigter,
Jack Collins,
Ingmar Posner %
}

\thanks{Applied AI Lab, Oxford Robotics Institute, University of Oxford}
\thanks{Correspondence to: {\tt\small jyamada@robots.ox.ac.uk}}%
}

\begin{document}

\maketitle
\thispagestyle{empty}
\pagestyle{empty}

\setlength{\abovedisplayskip}{6pt}
\setlength{\belowdisplayskip}{6pt}

\begin{abstract}
Model-based RL is a promising approach for real-world robotics due to its improved sample efficiency and generalization capabilities compared to model-free RL. However, effective model-based RL solutions for vision-based real-world applications require bridging the sim-to-real gap for any world model learnt. 
Due to its significant computational cost, standard domain randomisation does not provide an effective solution to this problem. This paper proposes \textit{TWIST} (Teacher-Student World Model Distillation for Sim-to-Real Transfer) to achieve efficient sim-to-real transfer of vision-based model-based RL using distillation.
Specifically, \textit{TWIST} leverages state observations as readily accessible, privileged information commonly garnered from a simulator to significantly accelerate sim-to-real transfer. Specifically, a teacher world model is trained efficiently on state information. At the same time, a matching dataset is collected of domain-randomised image observations.
The teacher world model then supervises a student world model that takes the domain-randomised image observations as input. By distilling the learned latent dynamics model from the teacher to the student model, \textit{TWIST} achieves efficient and effective sim-to-real transfer for vision-based model-based RL tasks. Experiments in simulated and real robotics tasks demonstrate that our approach outperforms naive domain randomisation and model-free methods in terms of sample efficiency and task performance of sim-to-real transfer.

\end{abstract}

\section{Introduction}
Deep reinforcement learning (RL) has successfully been applied to challenging control problems such as dexterous manipulation \cite{akkaya2019solving}, locomotion~\cite{haarnoja2018learning}, and Atari~\cite{mnih2015human}.
A particularly promising approach is~\emph{model-based} RL, which learns a \emph{world model} of the environment, and utilises this model for planning or policy optimisation.
Compared to \emph{model-free} approaches, model-based RL holds the potential for broader generalisation~\cite{yu2021combo}, improved sample efficiency~\cite{chua2018deep, deisenroth2011pilco}, and faster adaptation to new tasks~\cite{sekar2020planning, rigter2023reward}.
However, while model-based RL algorithms have been highly successful in simulated environments~\cite{hafner2022mastering, schrittwieser2020mastering}, their application to real-world robots remains limited due to the need for unsafe or costly data collection~\cite{levine2020offline} to train a world model in the real world.

Instead of training an RL agent directly in the real world, \textit{sim-to-real transfer} is a common approach: learning a policy from easily accessible simulated data and deploying it in the real environment.
In real-world environments, we often do not have access to accurate state information, and therefore we wish to learn a policy that utilises images as inputs.
To overcome the gap between the simulator and the real world, \emph{domain randomisation} (DR) is often employed.
DR exposes the policy to a wide range of simulated environments during training to improve generalisation to the real environment.
However, a significant drawback of DR is that policy training on randomised environments requires much more data~\cite{salter2021attention}.
Therefore, RL with DR can be extremely computationally intensive and may require weeks of computation time for training to converge~\cite{akkaya2019solving}.

The vast majority of existing work on sim-to-real transfer is applied to model-free RL~\cite{pinto2017asymmetric, brosseit2021distilled, salter2021attention, james2019simtoreal}.
In this work, we address the uninvestigated area of sim-to-real transfer for model-based RL trained from images.
By leveraging model-based RL algorithms, we benefit from the improved sample efficiency of model-based approaches~\cite{chua2018deep}.
However, to address the sim-to-real gap, it is still necessary to apply DR.
Similar to applying DR to the model-free case,
na\"ively applying DR to model-based approaches increases the amount of data required to train a suitable world model, and is therefore computationally very demanding~\cite{rigter2023reward}.

To address this, we propose \textbf{T}eacher-Student \textbf{W}orld model D\textbf{i}stillation for \textbf{S}im-to-Real \textbf{T}ransfer (\textit{TWIST}). 
\textit{TWIST} leverages privileged information in a simulator to achieve efficient and robust sim-to-real transfer for model-based RL.
In particular, \textit{TWIST} utilises two world models, a \textit{teacher} and a \textit{student}, to learn the environment.
The input to the \emph{teacher} is state information that is only accessible within the simulator. The teacher model is therefore unaffected by appearance changes as introduced by DR and can learn to represent the environment dynamics within a compact latent space much more efficiently than a vision-based model.
The teacher model then supervises a \emph{student} world model by encouraging it to encode domain-randomised image observations to the same latent representation as the teacher.
We demonstrate that \textit{TWIST} provides efficient and effective sim-to-real transfer for model-based RL, outperforming the standard DR-based approach almost by an order of magnitude in terms of success rate when applied to real-world manipulation tasks.

Our general approach of combining world model distillation with DR is applicable to any model-based RL algorithm.
In our implementation, we specifically use the DreamerV2 model architecture~\cite{hafner2022mastering} to learn the world models and associated policies, and apply our approach to a set of simulated and real robotics environments.
We show that our approach successfully achieves transfer to real-world environments, and outperforms na\"ive DR and model-free approaches in terms of sample efficiency and performance.
Our work demonstrates empirically, that there is significant potential for 
sim-to-real transfer of model-based RL, extending its applicability to a wide range of real-world robotics applications.

\section{Related Works}
The key concepts that \emph{TWIST} builds upon include model-based RL, sim-to-real transfer, and distillation using privileged information. 
We review the relevant literature of each of these concepts in turn.

\textbf{Model-based RL} has emerged as a promising approach to solving complex control problems by leveraging a learned dynamics model~\cite{ha2018world, hafner2019learning, sutton1991dyna}.
To achieve the desired behaviour, the dynamics model (or ``world'' model) can be used for planning~\cite{chua2018deep, RUBINSTEIN199789, schrittwieser2020mastering,  williams2015model}, or policy optimisation~\cite{ha2018world, hafner2022mastering, hafner2023mastering, rigter2022rambo}.
To handle partially observable environments~\cite{kaelbling1998planning} with high-dimensional observations such as images, a common approach is to employ a recurrent state-space model (RSSM)~\cite{schmidhuber1990reinforcement, ha2018world, hafner2019learning}, which predicts transitions in a compact latent space with a recurrent module. 
Despite considerable success on simulated environments, such as Atari~\cite{mnih2013playing} and DMControl~\cite{tassa2018deepmind}, applications of vision-based model-based RL to real-world robotics tasks remain limited due to the need for a large number of samples to train the world model~\cite{wu2022daydreamer, seo2023multi}.
Existing works on model-based RL from images for robotics~\cite{wu2022daydreamer, seo2023multi} build upon a suite of  
Dreamer algorithms~\cite{Hafner2020Dream, hafner2022mastering, hafner2023mastering}, which achieves state-of-the-art performance on simulated domains by optimising a policy using only synthetic data generated by the model.
DayDreamer~\cite{wu2022daydreamer} relies upon either state information or discretised action-spaces to simplify robotics tasks, and to facilitate learning a model from data collected directly in the real world.
Existing approaches to transferring Dreamer from simulation to real robots either require state information~\cite{brunnbauer2022latent} or only demonstrate transfer to near-identical real-world environments~\cite{seo2023multi}.

\textbf{Sim-to-real transfer}~\cite{zhao2020sim} trains a policy using simulated data, and deploys the policy in the real world.
Existing approaches to sim-to-real transfer utilise techniques such as domain randomisation (DR)~\cite{tobin2017domain}, system identification~\cite{Lutter2021DifferentiableLearning}, and domain adaptation~\cite{bousmalis2018using}.
DR is a particularly simple, yet effective approach to expose agents to a wide range of instances of the same environment by randomising visual and dynamics parameters.
By training policies using DR, agents become more robust to domain mismatches~\cite{tobin2017domain}.
Previous work on sim-to-real transfer using DR has been primarily applied to model-free RL methods~\cite{pinto2017asymmetric, salter2020attention, andrychowicz2020learning} or imitation learning~\cite{james2017transferring}.

Compared to sim-to-real transfer of model-free RL algorithms, model-based sim-to-real methods remain relatively unexplored. To our knowledge, \cite{brunnbauer2022latent} is the only work to transfer a model-based method across the sim-to-real gap. The authors accomplish this using a state-based Dreamer model that requires privileged information \textit{in the real world}.
Enabling sim-to-real transfer of Dreamer from image observations will help to unleash the potential of model-based RL for real-world applications where state information is not available.

Leveraging \textbf{privileged information} to accelerate the training of policies is a common approach.
Specifically, \cite{pinto2017asymmetric, salter2020attention} utilise information asymmetric actor-critic methods to train the critic faster via access to the privileged information while providing only images for the actor.

Another common technique to make use of the privilege information is \textbf{Distillation}, which transfers knowledge about a task from one or multiple teachers to a student.
In RL, knowledge transfer is generally achieved via \textit{policy} distillation: training a student policy to imitate a teacher policy~\cite{liu2022distilling, brosseit2021distilled, rusu2016policy, czarnecki2019distilling, chen2022system}.
Our work is most closely related in spirit to \cite{liu2022distilling, brosseit2021distilled, chen2022system} in that distillation and DR are used to efficiently train a teacher policy from privileged information and distil it into a student policy for sim-to-real transfer. 
However, for distillation, the prior works focus on model-free RL, which often requires additional trajectories collected by either the teacher or student policy to match the action distribution.

In contrast to these works, we consider model-based RL conditioned on image observations and introduce a novel method for \textit{world model} distillation. 
Our approach achieves knowledge transfer by supervising a student world model instead of a policy without the need for additional data collection during the distillation.
We demonstrate that our approach achieves strong performance for sim-to-real transfer in both simulated and real environments.

\begin{figure*}[t]
    \centering
    \includegraphics[width=0.9\textwidth]{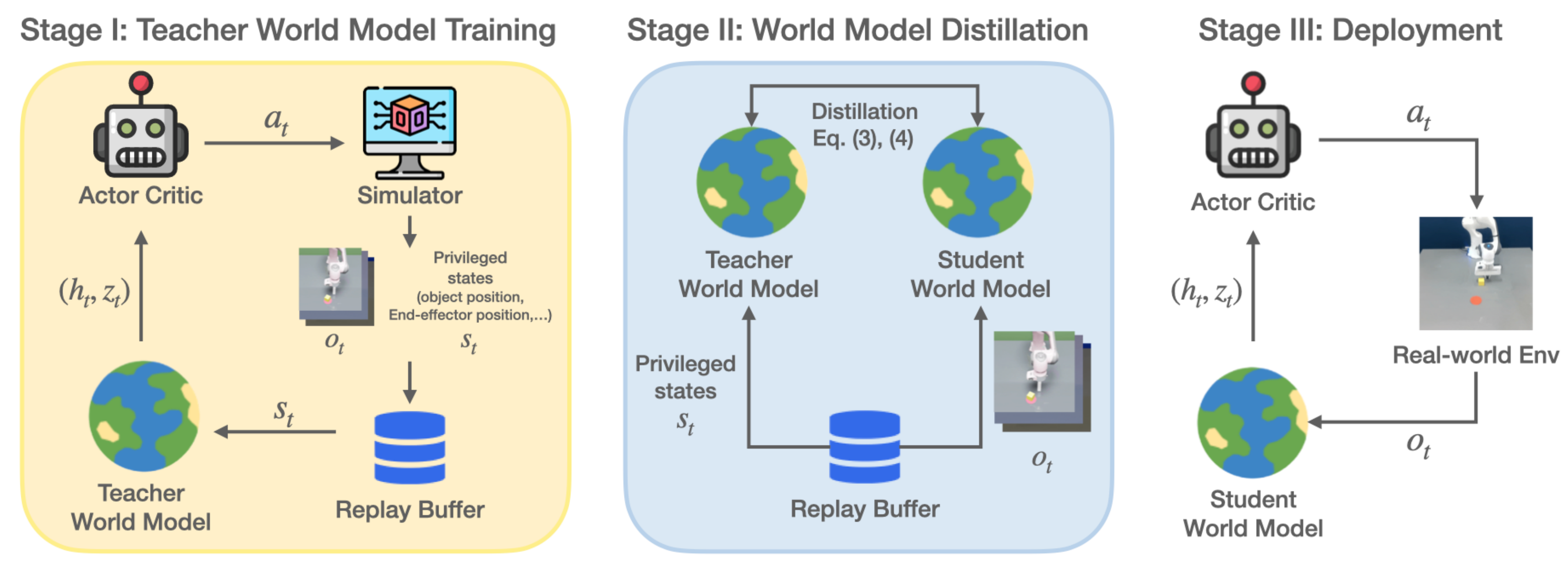}
    \caption{Overview of \textit{TWIST}. While a teacher world model is trained from privileged information, domain-randomised image observations are collected for distillation. The teacher supervises a student trained from the domain-randomised images to imitate the compact latent states of the teacher. The student world model is then transferred to real-world environments.}
    \label{method_overview}
    \vspace{-0.4cm}
\end{figure*}

\section{Preliminaries}
In this section, we describe our problem setting and the Dreamer model-based RL algorithm~\cite{Hafner2020Dream, hafner2022mastering, hafner2023mastering}.
We implement our approach using Dreamer as it is a commonly used state-of-the-art model-based RL algorithm that demonstrates the capability of our world model distillation approach.

\subsection{Problem Formulation}
The real environment is a partially observable Markov Decision Process represented by the tuple $(\mathcal{S}, \mathcal{O}, \mathcal{A}, \mathcal{P}, \mathcal{I}, r, \gamma, \mathbb{S})$, where: $\mathcal{S}$ is a set of continuous states, $\mathcal{O}$ is a set of image observations, $\mathcal{A}$ is a set of continuous actions, $\mathcal{P}:\mathcal{S} \times \mathcal{A} \times \mathcal{S}\rightarrow \mathbb{R}$ is the transition function, $\mathcal{I}: \mathcal{O} \times \mathcal{S} \rightarrow \mathbb{R}$ is the observation function, $r: \mathcal{S} \times \mathcal{A} \rightarrow \mathbb{R}$ is the reward function, $\gamma$ is the discount factor, and $\mathbb{S}$ is the initial state distribution.
The goal is to maximise the expected discounted reward $\mathbb{E}\left[\sum_{t=0}^{\infty} \gamma^t r_{t}\right]$.

In our problem setting, we do not have access to the real environment during training. Instead, we have access to a simulator that approximates the real environment.
In the simulator, we have direct access to privileged states, $s \in \mathcal{S}$, in addition to randomised image observations $o \in O$.

\subsection{Dreamer}

 Dreamer~\cite{hafner2022mastering, hafner2023mastering} is a model-based RL method that learns a world model from pixels or state observations and trains an actor-critic agent by leveraging imagined trajectories from the world model.

\paragraph{World Model} Dreamer uses a Recurrent State Space Model (RSSM)~\cite{hafner2019learning} to learn the dynamics of environments, consisting of the following modules:

\begin{equation}
\text { RSSM } \begin{cases}\text { Sequence model: } & h_t=f_\phi\left(h_{t-1}, z_{t-1}, a_{t-1}\right) \\ \text { Representation model: } & z_t \sim q_\phi\left(z_t \mid h_t, x_t\right) \\ \text { Dynamics predictor: } & \hat{z}_t \sim p_\phi\left(\hat{z}_t \mid h_t\right) \\ \text { Reward predictor: } & \hat{r}_t \sim p_\phi\left(\hat{r}_t \mid h_t, z_t\right) \\ \text { Decoder: } & \hat{x}_t \sim p_\phi\left(\hat{x}_t \mid h_t, z_t\right)\end{cases}
\end{equation}

All modules are implemented as neural networks parameterised by $\phi$.
In the RSSM, the state is jointly represented by a recurrent deterministic component, $h_t$, and a stochastic component represented by a categorical distribution.
At each step, the RSSM uses $h_t$ to compute two distributions over the stochastic state: $z_t$ and $\hat{z}$.
The stochastic posterior state $z_t$ encodes information about the current input observation $x_t$, while the prior state $\hat{z}_t$ is a prediction of the posterior state $z_t$ without access to the current input observation.
Therefore, by learning to predict $\hat{z}_t$, the model learns to predict the dynamics of the environment.
Given the posterior state, the decoder and reward predictor are trained to reconstruct the current input observation $x_t$ and the reward $r_t$, respectively. 
These models are jointly learned by minimising the negative variational lower bound~\cite{kingma2013auto}.
\begin{equation}
\begin{aligned}
\mathcal{L}(\theta) & \doteq \mathbb{E}_{q_\theta(s_{1: T} \mid a_{1: T}, x_{1: T})}\Big[\sum_{t=1}^T(-\ln p_\theta(x_t \mid h_t, z_t ) \\
& -\ln p_\theta(r_t \mid h_t, z_t) +\beta KL\left[q_\theta(z_t \mid h_{t}, x_{t}) \| p_\theta(\hat{z}_t \mid h_{t})\right]\Big]
\end{aligned}
\end{equation}

Once the model has been trained, it can be rolled out without access to any input observations by utilising the prior $\hat{z}$ in place of the posterior $z$.
This enables the model to generate unlimited synthetic or \emph{imagined} trajectories of the form: 
$\{\{h_t,\hat{z}_t,a_t,r_t\}_{t=0}^{t=T}\}$, where $T$ is the time horizon for imagination.

\paragraph{Actor-Critic Learning}
To learn a policy, Dreamer leverages an actor-critic algorithm that is trained using synthetic data generated by the world model.
Given a particular RSSM state $(h_t,\hat{z}_t)$, the critic is trained to predict the total expected reward.
The actor (i.e. the policy) is trained to output a distribution over actions, $\pi(a_{t}|h_{t}, \hat{z})$, that maximises the total expected reward given the current state.

\section{TWIST: Teacher-Student World Model Distillation for Sim-to-Real Transfer}
Dreamer is capable of efficiently solving diverse vision-based continuous control tasks in simulated environments by explicitly learning a task-agnostic world model. 
To transfer Dreamer to real-world robotics tasks, domain randomisation (DR) is required to bridge the gap between simulation and real-world environments.
However, DR dramatically increases the number of samples, and therefore computation time, required for training.
To address this issue, we propose \emph{TWIST} (\textbf{T}eacher-Student \textbf{W}orld Model D\textbf{i}stillation for \textbf{S}im-to-Real \textbf{T}ransfer)
to efficiently train a world model for vision-based tasks in simulation which readily transfers into real-world environments.
In this section, we describe our approach to distilling the teacher to the student world model (see Fig.~\ref{method_overview}).

\subsection{Overview}
A simulator affords access to state information in addition to domain-randomised images. 
\textit{TWIST} leverages this privileged information in order to accelerate the sim-to-real transfer of model-based RL.
Specifically, \textit{TWIST} initially trains a teacher world model and associated policy based on state information.
Because the teacher learns from state information, an accurate world model and strong policy can be trained from only a small number of samples.  

However, in real-world environments, privileged information is not usually available.
To overcome this issue, the teacher is distilled into a vision-based student world model.
While training the teacher from the state observations $s_{t}$, privileged information easily accessible in simulation, a matching dataset of domain-randomised image observations $o_t$ is generated, denoted as $\mathcal{D} = \{(s_{t}, o_{t}, a_{t}, r_{t}),...\}$.
The student is trained to imitate the RSSM latent states of the teacher while operating on the corresponding domain-randomised raw pixel inputs $o_{t}$ from the dataset $\mathcal{D}$.
Aligning these representations enables effective knowledge transfer and achieves sample-efficient sim-to-real transfer.

\subsection{World Model Distillation}
Given the teacher world model trained on state information, the teacher supervises the student to imitate the dynamics of the environment.
Specifically, the student is trained to imitate the prior distribution $p(\hat{z}_{t}^{\mathrm{teacher}} | h_{t}^{\mathrm{teacher}})$, posterior distribution $q(z_{t}^{\mathrm{teacher}} | h_{t}^{\mathrm{teacher}}, s_{t})$, and deterministic representations $h_{t}^{\mathrm{teacher}}$ of the teacher for a trajectory $\tau$ of length $L$ sampled from the dataset, $\mathcal{D}$:
\begin{equation}
\begin{array}{llll}
\end{array}
    \begin{aligned}
        &\mathcal{L}_{\mathrm{distil}}(\tau) = \mathbb{E}_{\{(a_{t}, o_{t}, s_{t})\}^{k+L}_{t=k}\sim \mathcal{D}} \sum_{t=k}^{k+L} \bigg[\underbrace{||h_{t}^{\mathrm{teacher}} - h_{t}^{\mathrm{student}}||^{2}_{2}}_{\text{Deterministic representation distillation}} \\
        &+ \underbrace{\mathbb{KL}[p_{\theta}(\hat{z}_{t}^{\mathrm{student}} | h_{t}^{\mathrm{student}}) || p_{\phi}(\hat{z}_{t}^{\mathrm{teacher}} | h_{t}^{\mathrm{teacher}})]}_{\text{Prior distillation}} \\
        &+ \underbrace{\mathbb{KL}[q_{\theta}(z_{t}^{\mathrm{student}} | h_{t}^{\mathrm{student}}, o_{t}) || q_{\phi}(z_{t}^{\mathrm{teacher}} | h_{t}^{\mathrm{teacher}}, s_{t})]}_{\text{Posterior distillation}}\bigg]
    \end{aligned}
    \label{eq:distill_loss}
\end{equation}
where $\phi$ and $\theta$ represent the parameters of the teacher and student 
world model, respectively.
Note that the parameter of the teacher world model $\phi$ is frozen during the distillation.

In addition to distilling the two stochastic distributions and deterministic representations, we further derive a training signal for distribution alignment by matching imagined rollouts in both the teacher and the student models. (Algorithm~\ref{alg:world_model_distillation}).
Specifically, a set of initial latent states in each world model is computed by embedding the trajectories $\tau$ sampled from the dataset $\mathcal{D}$ (see lines~\ref{line:embed1} and~\ref{line:embed2}).
Starting from the initial states of the teacher, we then generate an imagined rollout $\hat{\tau}^{\mathrm{teacher}} = \{(\hat{z}_{i}^{\mathrm{teacher}}, h_{i}^{\mathrm{teacher}}, a_{i}^{\mathrm{teacher}})\}_{i=t}^{t+H}$ with the time horizon $H$ using the policy $\pi$ learned with the teacher model (line \ref{line:imag1}).
We also collect an imagined trajectory $\tau^{student}$ in the student world model by replaying the same sequence of actions $\{a_{i}^{\mathrm{teacher}}\}_{i=1}^{H}$ used for trajectory imagination in the teacher (line~\ref{line:imag2}).
Then, we align the prior distribution $p(\hat{z}_{t} | h_{t})$ and deterministic representation $h_{t}$ in the trajectories generated by the teacher and student world model:
\begin{equation}
\begin{array}{llll}
\end{array}
    \begin{aligned}
        &\mathcal{L}_{\text{imagined}}(\hat{\tau}^{student}, \hat{\tau}^{\mathrm{teacher}}) =  \sum_{i=1}^{H} \bigg[ \underbrace{||h_{i}^{\mathrm{teacher}} - h_{i}^{student}||^{2}_{2}}_{\text{Deterministic representation distillation}} \\
        &+ \underbrace{\mathbb{KL}[p_{\theta}(\hat{z}_{i}^{\mathrm{student}} | h_{i}^{\mathrm{student}}) || p_{\phi}(\hat{z}_{i}^{\mathrm{teacher}} | h_{i}^{\mathrm{teacher}})]}_{\text{Distillation in Imagination}}\bigg]
    \end{aligned}
    \label{eq:imagined_distill}
\end{equation}
where $(h_i^{\mathrm{teacher}}, z_i^{\mathrm{teacher}})$ and $(h_i^{\mathrm{student}}, z_i^{\mathrm{student}})$ are the $i^{th}$ entries in $\hat{\tau}^{\mathrm{teacher}}$ and $\hat{\tau}^{\mathrm{student}}$ respectively.
To ensure diversity in the imagined trajectories, random noise is added to the action $a_{t}$ sampled from the policy $\pi(a_{t}|h_{t}^{\mathrm{teacher}}, \hat{z}_{t}^{\mathrm{teacher}})$ when rolling it out in the teacher world model.
This bootstraps the trajectories in the dataset $\mathcal{D}$; thus, the student can imitate the prior distribution and deterministic representation of the teacher more accurately.
The loss function for world model distillation is therefore $ \mathcal{L} = \mathcal{L}_{\text{distill}} + \mathcal{L}_{\text{imagined}}$.
Our experimental results demonstrate that, after distillation, an actor trained in the teacher world model successfully transfers to real-world environments as the student world model is trained to imitate the RSSM latent states of the teacher.

\begin{algorithm}[t]
    \caption{TWIST: Teacher-Student World Model Distillation for Sim-To-Real Transfer}
    \label{alg:world_model_distillation}
    \begin{algorithmic}[1]
    \State \textbf{Inputs:} Dataset $\mathcal{D} = \{(s_{i}, o_{t}, a_{t}, r_{t}),...\}$; Teacher world model $W^{\mathrm{teacher}}_{\phi}$; Policy $\pi(a_{t} | h_{t}, \hat{z}_{t})$
    \State \textbf{Initialise:} Student world model $W^{\mathrm{student}}_{\theta}$
    \While{distilling world model}
    \State $\tau = \{(a_{t}, o_{t}, s_{t})\}_{t=k}^{k+L} \sim D$ \label{line:sample}
    \State Compute $\mathcal{L}_{\mathrm{distill}}$ via Eq.~\ref{eq:distill_loss} using $\tau$ \label{line:standard_distil}
    \State  $\mathcal{Z}_{\tau}^{\mathrm{teacher}} = \{z_t^{\mathrm{teacher}}\}_{t=k}^{k+L} \leftarrow q_{\phi}(\tau)$ \label{line:embed1}
    \State  $\mathcal{Z}_{\tau}^{\mathrm{student}} = \{z_t^{\mathrm{student}}\}_{t=k}^{k+L} \leftarrow q_{\theta}(\tau)$ \label{line:embed2}
    \State $\hat{\tau}^{teacher}$ = \Call{Imagine}{$W_{\phi}^{\mathrm{teacher}}$, $Z_{\tau}^{\mathrm{teacher}}$} \label{line:imag1}
    \State $A^{\mathrm{teacher}} \leftarrow \{a_{i}\}_{i=1}^{H}$ in $\hat{\tau}^{\mathrm{teacher}}$
    \State $\hat{\tau}^{\mathrm{student}}$ = \Call{Imagine}{$W_{\theta}^{\mathrm{student}}$, $Z_{\tau}^{\mathrm{student}}$, $A^{\mathrm{teacher}}$}\label{line:imag2}
    \State Compute $\mathcal{L}_{\mathrm{imagined}}$ via Eq.~\ref{eq:imagined_distill}
    \State $\theta \leftarrow \theta - \alpha \nabla_{\theta} (\mathcal{L}_{\mathrm{distill}} + \mathcal{L}_{\mathrm{imagined}})$ 
    \EndWhile \vspace{1mm}
    \Function{Imagine}{$W$, $Z_{\mathrm{init}}$, $A=\text{None}$}
        \If{$A$ is None} \Comment{Imagination in $W^{\mathrm{teacher}}$}
            \State $\hat{\tau} \leftarrow$ rollout $\pi$ for H steps from $z \in Z_{\mathrm{init}}$ in $W$ 
        \Else  \Comment{Imagination in $W^{\mathrm{student}}$}
            \State $\hat{\tau} \leftarrow$ rollout $a \in A$ from $z \in Z$ in $W$
        \EndIf
        \State \Return $\hat{\tau}$ \Comment{$\hat{\tau} = \{(\hat{z}_{i}, h_i, a_{i})\}_{i=1}^{H}$}
    \EndFunction
    \end{algorithmic}    
\end{algorithm}

\section{Implementation Details}
Our encoder and decoder of the teacher world model consists of three fully connected hidden layers with $512$ units and ELU activation.
We use the same architecture for the encoder, decoder, and actor-critic agent of vision-based world models as those used in~\cite{hafner2022mastering}.
For distillation, a trajectory of length $L=50$ is sampled from the dataset $\mathcal{D}$ (see Eq.~\ref{eq:distill_loss}) and an imagined trajectory of length $H=15$ is generated (see Eq.~\ref{eq:imagined_distill}).
All of the agents are trained on a single GeForce RTX 3090 for $500K$ environment steps.

\begin{table*}[ht]
    \centering
    \begin{tabular}{l || c | c  c c  c  c  c}
    \hline
      Tasks (500K Steps) & Oracle Dreamer & TWIST & Dreamer w/o DR & Dreamer w/ DR & Dreamer State Recon. & Asymmetric SAC \\
     \hline
    Cup Catch  & $936.6 \pm 0.1$ & \cellcolor{blue!10} $856.6 \pm 29.6$ & $150.7 \pm 80.3$ & $744.3 \pm 93.8$ & $627.0 \pm 194.7$ & \cellcolor{blue!10}$873.0 \pm 11.1$  \\ 
     
     Cartpole, Balance & $992.9 \pm 1.7$ & \cellcolor{blue!10} $954.5 \pm 37.4$  & $349.3 \pm 19.0$ & $590.8 \pm 23.0$  & $869.7 \pm 40.4$ & $353.1 \pm 18.6$   \\
     
     Cheetah Run & $597.1 \pm 24.3$ & \cellcolor{blue!10} $506.0 \pm 54.2 $ & $206.5 \pm 40.3$ & $476.4 \pm 61.1$ & $391.6 \pm 4.5$ & $222.5 \pm 20.4$   \\ 
     Hopper Stand  &  $501.1 \pm 38.9 $ & \cellcolor{blue!10} $483.3 \pm 118.9$ & $42.1 \pm 11.2$ & \cellcolor{blue!10} $471.8 \pm 48.4$ & $358.2 \pm 34.4$ & $57.7 \pm 86.8$ \\      
     Walker Walk  & $800.4 \pm 53.7$  & \cellcolor{blue!10} $665.8 \pm 80.3$ & $182.7 \pm 3.4$ & $394.6 \pm 25.1$ & $491.0 \pm 145.6$ & $439.4 \pm 57.0$    \\      
     Finger, Easy Turn & $904.4 \pm 32.8$  & \cellcolor{blue!10} $798.0 \pm 50.3$ & $182.5 \pm 26.0$ & $440.7 \pm 35.7$ & $553.4 \pm 42.2$ & $304.4 \pm 22.5$    \\      
     
     \hline
    \end{tabular}
    \caption{Averaged episodic rewards and standard deviation obtained from 100 trials with 3 seeds in the Distracting Control Suite. The evaluation is conducted using held-out environments.
}
    \label{table:success_results}
    \vspace{-0.25cm}
\end{table*}

\section{Experiments}

The efficacy of \emph{TWIST} for sim-to-real transfer is evaluated through experiments in both simulated and real-world environments. 
The experiments aim to answer the following questions: (1) does \emph{TWIST} enable efficient sim-to-real transfer for model-based RL using DR? and (2) does the distillation for imagined trajectories improve the task performance compared to performing distillation only on the original dataset?

\subsection{Baselines}
We compare \emph{TWIST} against several competitive baselines, including Dreamer agents with different training methods and model-free RL.
\textit{Oracle} is a Dreamer agent trained from privileged information. The performance of the oracle agent is an upper bound on the performance of our method. Since we do not have access to state information in real-world settings, we only provide the performance of the oracle approach in the experiments conducted in simulation environments.
\textit{Dreamer w/ DR} is a vision-based Dreamer agent trained with naive DR. \textit{Dreamer w/o DR} is an agent trained without DR.
\textit{Dreamer State Recon.} is a vision-based Dreamer agent trained to reconstruct state information from domain-randomised image observations, which is an alternative way of leveraging privileged information.
Lastly, \textit{Asymmetric SAC}~\cite{pinto2017asymmetric} is a sample-efficient state-of-the-art model-free RL algorithm suitable for DR. While the critic network is trained from privileged information, the policy is trained from domain-randomised image observations.

\begin{figure}[ht]
    \centering
   \includegraphics[width=0.85\linewidth]{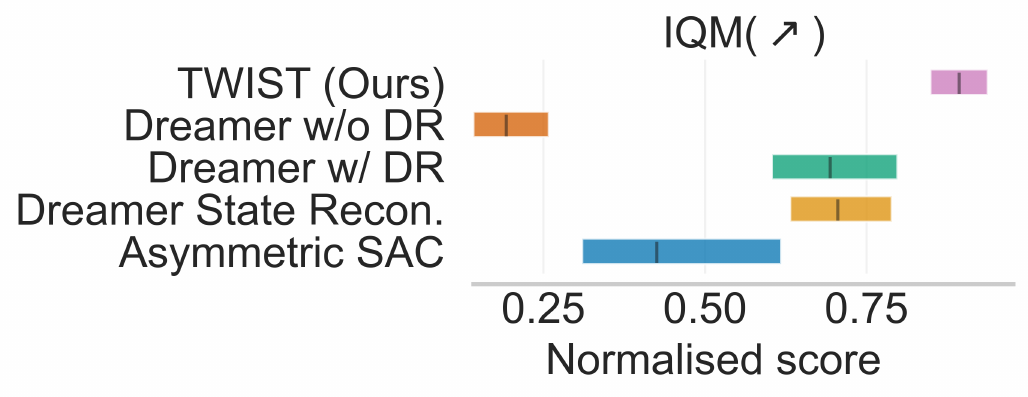}
    \caption{
    Aggregated Interquartile Mean (IQM) of normalised episodic rewards with $95\%$ bootstrap CI based on $5$ tasks from 100 trials with $3$ seeds evaluated using held-out environments in Distracting Control.
    The lack of overlap with the CIs between TWIST and the baseline methods indicates that the difference is statistically significant.
    }
\label{fig:main_stats}
\vspace{-0.2cm}
\end{figure}

\subsection{Simulated Results}
Firstly, we empirically demonstrate the efficacy of \emph{TWIST} on a set of continuous control tasks in the Distracting Control Suite~\cite{stone2021distracting}, an extended version of the DMControl~\cite{tassa2018deepmind}. 

\subsubsection{Experiment Setup}
First, a teacher world model and a policy are trained using Dreamer from ground-truth state information. During training, domain-randomised images are collected by randomizing the background texture used in prior work~\cite{hansen2021generalization} and the colour of objects every timestep for diverse data acquisition.
 After training the teacher, we use the domain-randomised image observations to distil the state-based teacher world model into a vision-based student world model.
For evaluation, we sample the object colours from the same distribution as training, but the background texture is sampled from a held-out test distribution. Therefore, the distribution of environments for evaluation is different to the training time environments.
Note that DR is applied only at the beginning of the episode for the evaluation because the textures are usually consistent at test time.

\subsubsection{Results}
Table ~\ref{table:success_results} reports the average episodic rewards for six continuous control tasks evaluated on hold-out scenes from the Distracting Control Suite.
\emph{TWIST} outperforms the baseline approaches, including model-free RL, often by significant margins.
While \textit{Asymmetric SAC} shows comparable performance on the simple \textit{Cup Catch} task, it does not perform well on more complex tasks because the policy struggles to learn task-relevant information efficiently from domain-randomised images due to its visual complexity.
\textit{Dreamer State Recon.}  and \textit{Dreamer w/ DR} demonstrate better performance among the baselines. 
However, learning task-relevant information and the actor-critic agent jointly on limited samples is often challenging, resulting in worse performance compared to our approach.
\textit{Dreamer w/o DR} does not perform well in any of the six tasks due to the lack of generalisation to unseen scenes.

To assess the statistical significance of our results, Fig.~\ref{fig:main_stats} reports Interquartile Mean (IQM) of normalised episodic rewards with $95\%$ bootstrap confidence interval (CI) aggregated across $5$ tasks in Distracting Control, computed using~\cite{agarwal2021deep}.
The episodic rewards of each task are normalised by the performance of \textit{Oracle Dreamer} to aggregate the results and validate the efficacy of our method.
As shown in Fig.~\ref{fig:main_stats}, our method is substantially more performant than the baselines. The lack of overlap with the CIs of the baseline method further indicates that this difference is statistically significant.

\begin{figure}[t]
    \centering
    \includegraphics[width=0.45\textwidth]{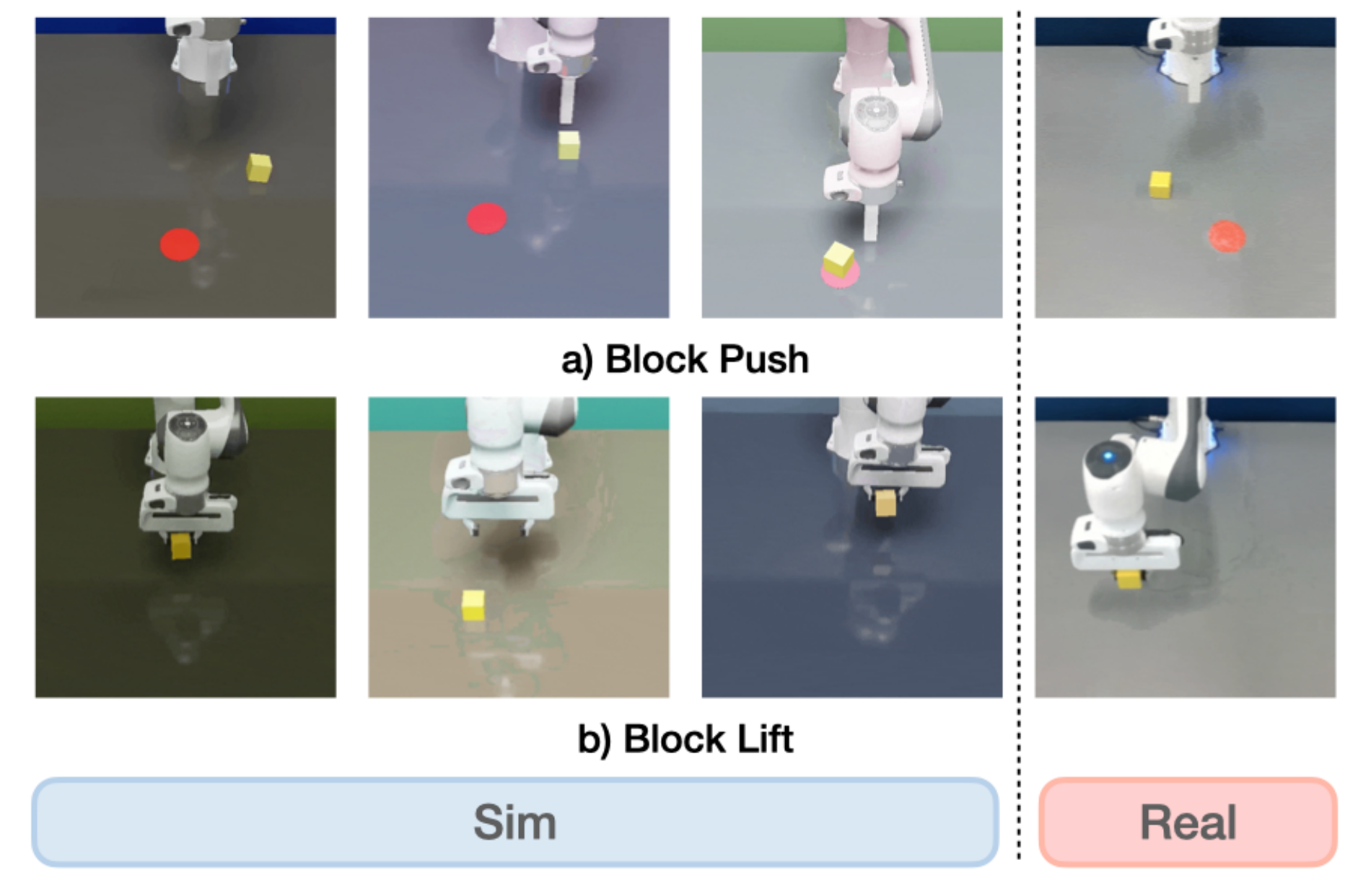}
    \caption{Sim-to-real manipulation tasks. (a) \textit{Block Push}: A Franka Panda arm pushes the yellow block towards the red goal marker. (b) \textit{Block Lift}: The arm grasps the yellow block and lifts it $10 \mathrm{cm}$ above the table} 
    \label{fig:tasks}
    \vspace{-0.3cm}
\end{figure}

\begin{figure*}
    \centering
    \includegraphics[width=1\textwidth]{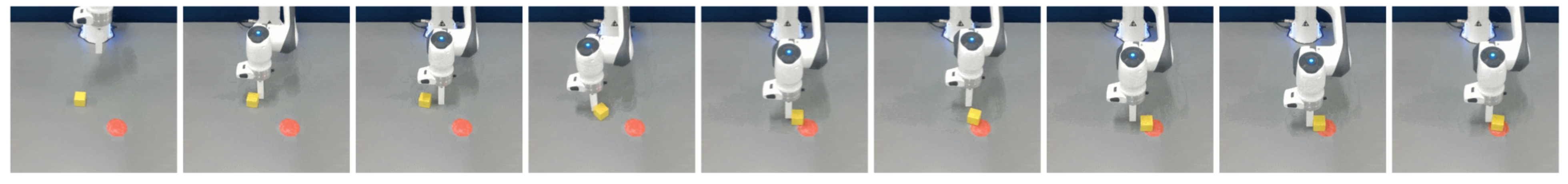}
    \label{fig:push}
    \vspace{-0.3cm}
    \caption{Example rollouts of the proposed method on the real-world Block Push task. Our method successfully transfers the student world model and solves the block push task in the real world.}
    \vspace{-0.3cm}
\end{figure*}

\subsection{Sim-to-Real Transfer for Manipulation Tasks}
In this section, we consider sim-to-real transfer for manipulation tasks to verify the effectiveness of \textit{TWIST} in the real world.

\subsubsection{Experimental setup}
In our experiments, a Franka Panda robot (7-DoF robot arm) is used.
In real-world experiments, RGB image observations are taken from a RealSense D435i camera. The MoveIt library~\cite{coleman2014reducing} is used to control the end-effector position.
In the simulation, agents are trained in Omniverse Isaac Orbit~\cite{mittal2023orbit} powered by Omniverse Isaac Sim~\cite{IsaacDeveloper}.
DR is applied to the brightness of the light and texture of the robot body, background, table, and objects every timestep to collect diverse image observations.
Further, the friction of objects is randomised in every episode.
The action space of the policy is a delta-position of the end-effector in Cartesian coordinates with a maximum delta of $2\mathrm{cm}$.

\begin{figure}[t]
    \centering
    \includegraphics[width=0.41\textwidth]{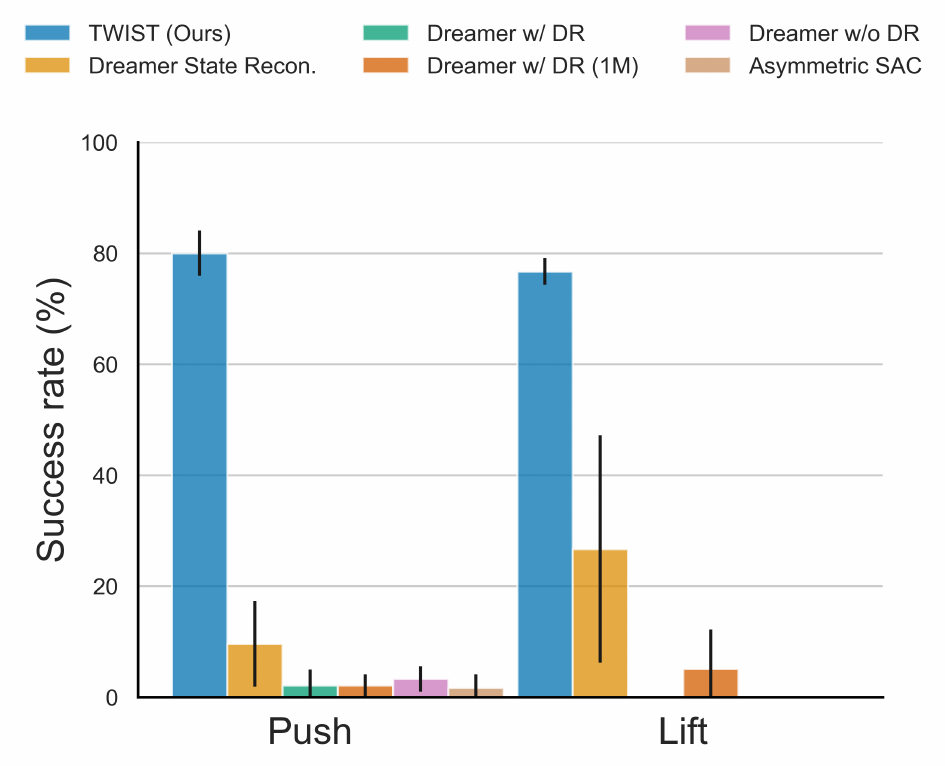}
    \vspace*{-3mm}
    \caption{Success rate on real-world tasks. The success rate and standard deviation are calculated from 20 trials with 3 seeds. \textit{TWIST} significantly outperform baselines including na\"ive Dreamer with DR and model-free RL.
    }
    \label{real_world}
    \vspace{-0.28cm}
\end{figure}

\subsubsection{Tasks}
We conduct experiments to showcase the successful sim-to-real transfer capability of \textit{TWIST}, focusing on the \textit{block push} and \textit{block lift} tasks (see Fig.~\ref{fig:tasks}).
The objective of the \textit{block push} task is to push a $4\mathrm{cm} \times 4\mathrm{cm}$ cube towards a designated red goal marker.
If the distance between the centre of the cube and the goal marker is less than $5\mathrm{cm}$ at the end of the episode, then the trial is considered successful.
The cube and goal marker positions are randomly sampled from a uniform distribution.
For the block push task, we replace the robot's hand with a 3D-printed peg to push the block because the original robot's hand often occludes the block from the third-person camera.

The goal of the \textit{block lift} task is to grasp the cube and lift it $10\mathrm{cm}$ above the tabletop by the end of the episode.
To train agents in simulation, we define a dense reward function tailored to each task.
The episode length of these tasks is 150 timesteps.
In real-world experiments, we randomise the camera position and brightness of the scene randomly to ensure robustness of the distilled Dreamer agents.

\subsubsection{Results}
The success rate for each task across $20$ trials averaged over $3$ seeds is reported in Fig.~\ref{real_world}.
Compared to the baselines, including na\"ive Dreamer with DR and model-free RL, \textit{TWIST} demonstrates significantly better success rates in both \textit{block push} and \textit{lift} tasks.
In particular, the block push task requires an accurate dynamics model to successfully push the box towards the goal marker, indicating that our world model is successfully distilled and transferred from simulation to real-world environments.
The baseline methods often fail to solve the task, because those methods require more samples to successfully train agents in simulation with DR~\cite{rigter2023reward}.
\textit{Dreamer State Recon.} shows a better success rate than other baselines.
However, it still struggles to learn task-relevant information in image observations effectively while exploring environments for solving manipulation tasks.
Although na\"ive Dreamer agent with DR is also trained from $1\mathrm{M}$ samples (\textit{Dreamer w/ DR ($1\mathrm{M}$)}), its success rate on the \textit{block push} and \textit{block lift} tasks remains low, indicating the sample inefficiency of the na\"ive DR approach.

\begin{figure}[ht]
    \centering
   \includegraphics[width=0.8\linewidth]{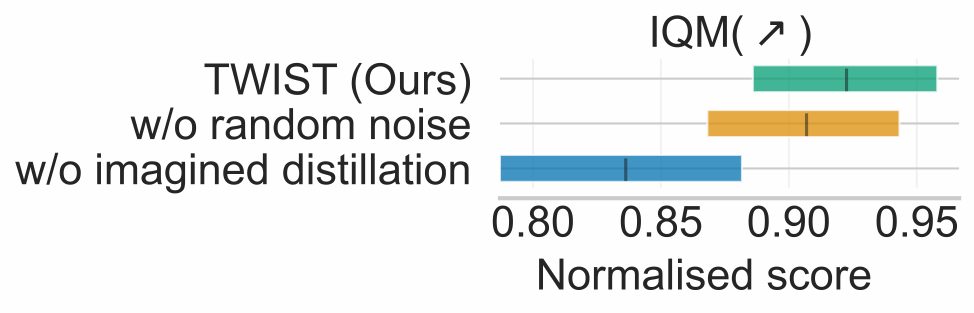}
   \label{fig:ablation_metrics}
    \caption{
    Interquartile Mean (IQM) of normalised episodic rewards with $95\%$ bootstrap CI to ablate the key components of the proposed distillation method in the Distracting Control Suite. The following variants are compared: (1) the full proposed method, (2) without random noise to actions for imagined distillation,  (3) without imagined distillation. 
    }
\label{fig:ablation_study}
\vspace{-0.25cm}
\end{figure}

\subsection{Ablation Study}
\label{sec:ablation}
We ablate the distillation for imagined trajectories (\textit{imagined distillation}) (see Eq.~\ref{eq:imagined_distill}) and random noise added to actions for the imagined distillation in the Distracting Control Suite.
We report normalised aggregated Interquartile Mean (IQM) with a $95\%$ bootstrap confidence interval.
As shown in Fig.~\ref{fig:ablation_study}, the CI for \textit{our method} and \textit{our method w/o imagined distillation} do not overlap, indicating that the difference in performance is statistically significant.
On the other hand, the gap between \textit{our method} and \textit{our method w/o random noise} is smaller but still notable in practice.
These results highlight that distillation using imagined rollouts is particularly important for successful world model distillation.

\section{Conclusion}
We propose \textit{TWIST} for efficient sim-to-real transfer of model-based RL.
Specifically, a teacher world model trained from privileged state information supervises a student world model taking as input domain-randomised image observations to mimic the compact latent states of the teacher.
Our experiments demonstrate successful distillation from the teacher world model to the student world model with domain randomisation in simulated environements and further show the efficient and robust sim-to-real transfer for robot manipulation tasks into real-world domains.

\textit{TWIST} is therefore a significant step towards unlocking the benefits of model-based RL for real-world applications.
In future work we will look to explore fine-tuning and reusing the distilled world model from few real-world image observations to efficiently acquire new skills in the real world.
\section*{ACKNOWLEDGMENT}
This work was supported by a UKRI/EPSRC Programme Grant [EP/V000748/1]. We would also like to thank the University of Oxford for providing Advanced Research Computing (ARC) and the SCAN facility in carrying out this work  (\url{http://dx.doi.org/10.5281/zenodo.22558}).

\clearpage
\bibliographystyle{IEEEtran}
\bibliography{bib/deep_learning, bib/distillation, bib/rl, bib/sim2real, bib/misc}

\begin{thebibliography}{10}
\providecommand{\url}[1]{#1}
\csname url@rmstyle\endcsname
\providecommand{\newblock}{\relax}
\providecommand{\bibinfo}[2]{#2}
\providecommand\BIBentrySTDinterwordspacing{\spaceskip=0pt\relax}
\providecommand\BIBentryALTinterwordstretchfactor{4}
\providecommand\BIBentryALTinterwordspacing{\spaceskip=\fontdimen2\font plus
\BIBentryALTinterwordstretchfactor\fontdimen3\font minus
  \fontdimen4\font\relax}
\providecommand\BIBforeignlanguage[2]{{%
\expandafter\ifx\csname l@#1\endcsname\relax
\typeout{** WARNING: IEEEtran.bst: No hyphenation pattern has been}%
\typeout{** loaded for the language `#1'. Using the pattern for}%
\typeout{** the default language instead.}%
\else
\language=\csname l@#1\endcsname
\fi
#2}}

\bibitem{akkaya2019solving}
I.~Akkaya, M.~Andrychowicz, M.~Chociej, M.~Litwin, B.~McGrew, A.~Petron,
  A.~Paino, M.~Plappert, G.~Powell, R.~Ribas, \emph{et~al.}, ``Solving rubik's
  cube with a robot hand,'' \emph{arXiv preprint arXiv:1910.07113}, 2019.

\bibitem{haarnoja2018learning}
T.~Haarnoja, S.~Ha, A.~Zhou, J.~Tan, G.~Tucker, and S.~Levine, ``Learning to
  walk via deep reinforcement learning,'' \emph{Robotics: Science and Systems},
  2019.

\bibitem{mnih2015human}
V.~Mnih, K.~Kavukcuoglu, D.~Silver, A.~A. Rusu, J.~Veness, M.~G. Bellemare,
  A.~Graves, M.~Riedmiller, A.~K. Fidjeland, G.~Ostrovski, \emph{et~al.},
  ``Human-level control through deep reinforcement learning,'' \emph{nature},
  vol. 518, no. 7540, pp. 529--533, 2015.

\bibitem{yu2021combo}
T.~Yu, A.~Kumar, R.~Rafailov, A.~Rajeswaran, S.~Levine, and C.~Finn, ``Combo:
  Conservative offline model-based policy optimization,'' \emph{Advances in
  neural information processing systems}, vol.~34, pp. 28\,954--28\,967, 2021.

\bibitem{chua2018deep}
K.~Chua, R.~Calandra, R.~McAllister, and S.~Levine, ``Deep reinforcement
  learning in a handful of trials using probabilistic dynamics models,''
  \emph{Advances in neural information processing systems}, vol.~31, 2018.

\bibitem{deisenroth2011pilco}
M.~Deisenroth and C.~E. Rasmussen, ``Pilco: A model-based and data-efficient
  approach to policy search,'' in \emph{Proceedings of the 28th International
  Conference on machine learning (ICML-11)}, 2011, pp. 465--472.

\bibitem{sekar2020planning}
R.~Sekar, O.~Rybkin, K.~Daniilidis, P.~Abbeel, D.~Hafner, and D.~Pathak,
  ``Planning to explore via self-supervised world models,'' in
  \emph{International Conference on Machine Learning}.\hskip 1em plus 0.5em
  minus 0.4em\relax PMLR, 2020, pp. 8583--8592.

\bibitem{rigter2023reward}
M.~Rigter, M.~Jiang, and I.~Posner, ``Reward-free curricula for training robust
  world models,'' \emph{arXiv preprint arXiv:2306.09205}, 2023.

\bibitem{hafner2022mastering}
D.~Hafner, T.~Lillicrap, M.~Norouzi, and J.~Ba, ``Mastering atari with discrete
  world models,'' 2022.

\bibitem{schrittwieser2020mastering}
J.~Schrittwieser, I.~Antonoglou, T.~Hubert, K.~Simonyan, L.~Sifre, S.~Schmitt,
  A.~Guez, E.~Lockhart, D.~Hassabis, T.~Graepel, \emph{et~al.}, ``Mastering
  atari, go, chess and shogi by planning with a learned model,'' \emph{Nature},
  vol. 588, no. 7839, pp. 604--609, 2020.

\bibitem{levine2020offline}
S.~Levine, A.~Kumar, G.~Tucker, and J.~Fu, ``Offline reinforcement learning:
  Tutorial, review, and perspectives on open problems,'' \emph{arXiv preprint
  arXiv:2005.01643}, 2020.

\bibitem{salter2021attention}
S.~Salter, D.~Rao, M.~Wulfmeier, R.~Hadsell, and I.~Posner,
  ``Attention-privileged reinforcement learning,'' in \emph{Conference on Robot
  Learning}.\hskip 1em plus 0.5em minus 0.4em\relax PMLR, 2021, pp. 394--408.

\bibitem{pinto2017asymmetric}
L.~Pinto, M.~Andrychowicz, P.~Welinder, W.~Zaremba, and P.~Abbeel, ``Asymmetric
  actor critic for image-based robot learning,'' 2017.

\bibitem{brosseit2021distilled}
J.~Brosseit, B.~Hahner, F.~Muratore, M.~Gienger, and J.~Peters, ``Distilled
  domain randomization,'' \emph{arXiv preprint arXiv:2112.03149}, 2021.

\bibitem{james2019simtoreal}
S.~James, P.~Wohlhart, M.~Kalakrishnan, D.~Kalashnikov, A.~Irpan, J.~Ibarz,
  S.~Levine, R.~Hadsell, and K.~Bousmalis, ``Sim-to-real via sim-to-sim:
  Data-efficient robotic grasping via randomized-to-canonical adaptation
  networks,'' 2019.

\bibitem{ha2018world}
D.~Ha and J.~Schmidhuber, ``World models,'' \emph{arXiv preprint
  arXiv:1803.10122}, 2018.

\bibitem{hafner2019learning}
D.~Hafner, T.~Lillicrap, I.~Fischer, R.~Villegas, D.~Ha, H.~Lee, and
  J.~Davidson, ``Learning latent dynamics for planning from pixels,'' in
  \emph{International conference on machine learning}.\hskip 1em plus 0.5em
  minus 0.4em\relax PMLR, 2019, pp. 2555--2565.

\bibitem{sutton1991dyna}
R.~S. Sutton, ``Dyna, an integrated architecture for learning, planning, and
  reacting,'' \emph{ACM Sigart Bulletin}, vol.~2, no.~4, pp. 160--163, 1991.

\bibitem{RUBINSTEIN199789}
\BIBentryALTinterwordspacing
R.~Y. Rubinstein, ``Optimization of computer simulation models with rare
  events,'' \emph{European Journal of Operational Research}, vol.~99, no.~1,
  pp. 89--112, 1997. [Online]. Available:
  \url{https://www.sciencedirect.com/science/article/pii/S0377221796003852}
\BIBentrySTDinterwordspacing

\bibitem{williams2015model}
G.~Williams, A.~Aldrich, and E.~Theodorou, ``Model predictive path integral
  control using covariance variable importance sampling,'' 2015.

\bibitem{hafner2023mastering}
D.~Hafner, J.~Pasukonis, J.~Ba, and T.~Lillicrap, ``Mastering diverse domains
  through world models,'' \emph{arXiv preprint arXiv:2301.04104}, 2023.

\bibitem{rigter2022rambo}
M.~Rigter, B.~Lacerda, and N.~Hawes, ``{RAMBO-RL}: Robust adversarial
  model-based offline reinforcement learning,'' \emph{Advances in Neural
  Information Processing Systems}, 2022.

\bibitem{kaelbling1998planning}
L.~P. Kaelbling, M.~L. Littman, and A.~R. Cassandra, ``Planning and acting in
  partially observable stochastic domains,'' \emph{Artificial intelligence},
  vol. 101, no. 1-2, pp. 99--134, 1998.

\bibitem{schmidhuber1990reinforcement}
J.~Schmidhuber, ``Reinforcement learning in markovian and non-markovian
  environments,'' \emph{Advances in neural information processing systems},
  vol.~3, 1990.

\bibitem{mnih2013playing}
V.~Mnih, K.~Kavukcuoglu, D.~Silver, A.~Graves, I.~Antonoglou, D.~Wierstra, and
  M.~Riedmiller, ``Playing atari with deep reinforcement learning,'' 2013.

\bibitem{tassa2018deepmind}
Y.~Tassa, Y.~Doron, A.~Muldal, T.~Erez, Y.~Li, D.~de~Las~Casas, D.~Budden,
  A.~Abdolmaleki, J.~Merel, A.~Lefrancq, T.~Lillicrap, and M.~Riedmiller,
  ``Deepmind control suite,'' 2018.

\bibitem{wu2022daydreamer}
P.~Wu, A.~Escontrela, D.~Hafner, K.~Goldberg, and P.~Abbeel, ``Daydreamer:
  World models for physical robot learning,'' 2022.

\bibitem{seo2023multi}
Y.~Seo, J.~Kim, S.~James, K.~Lee, J.~Shin, and P.~Abbeel, ``Multi-view masked
  world models for visual robotic manipulation,'' \emph{arXiv preprint
  arXiv:2302.02408}, 2023.

\bibitem{Hafner2020Dream}
\BIBentryALTinterwordspacing
D.~Hafner, T.~Lillicrap, J.~Ba, and M.~Norouzi, ``Dream to control: Learning
  behaviors by latent imagination,'' in \emph{International Conference on
  Learning Representations}, 2020. [Online]. Available:
  \url{https://openreview.net/forum?id=S1lOTC4tDS}
\BIBentrySTDinterwordspacing

\bibitem{brunnbauer2022latent}
A.~Brunnbauer, L.~Berducci, A.~Brandst{\'a}tter, M.~Lechner, R.~Hasani, D.~Rus,
  and R.~Grosu, ``Latent imagination facilitates zero-shot transfer in
  autonomous racing,'' in \emph{2022 International Conference on Robotics and
  Automation (ICRA)}.\hskip 1em plus 0.5em minus 0.4em\relax IEEE, 2022, pp.
  7513--7520.

\bibitem{zhao2020sim}
W.~Zhao, J.~P. Queralta, and T.~Westerlund, ``Sim-to-real transfer in deep
  reinforcement learning for robotics: a survey,'' in \emph{2020 IEEE symposium
  series on computational intelligence (SSCI)}.\hskip 1em plus 0.5em minus
  0.4em\relax IEEE, 2020, pp. 737--744.

\bibitem{tobin2017domain}
J.~Tobin, R.~Fong, A.~Ray, J.~Schneider, W.~Zaremba, and P.~Abbeel, ``Domain
  randomization for transferring deep neural networks from simulation to the
  real world,'' in \emph{2017 IEEE/RSJ international conference on intelligent
  robots and systems (IROS)}.\hskip 1em plus 0.5em minus 0.4em\relax IEEE,
  2017, pp. 23--30.

\bibitem{Lutter2021DifferentiableLearning}
M.~Lutter, J.~Silberbauer, J.~Watson, and J.~Peters, ``Differentiable physics
  models for real-world offline model-based reinforcement learning,'' in
  \emph{2021 IEEE International Conference on Robotics and Automation (ICRA)},
  2021, pp. 4163--4170.

\bibitem{bousmalis2018using}
K.~Bousmalis, A.~Irpan, P.~Wohlhart, Y.~Bai, M.~Kelcey, M.~Kalakrishnan,
  L.~Downs, J.~Ibarz, P.~Pastor, K.~Konolige, \emph{et~al.}, ``Using simulation
  and domain adaptation to improve efficiency of deep robotic grasping,'' in
  \emph{2018 IEEE international conference on robotics and automation
  (ICRA)}.\hskip 1em plus 0.5em minus 0.4em\relax IEEE, 2018, pp. 4243--4250.

\bibitem{salter2020attention}
\BIBentryALTinterwordspacing
S.~Salter, D.~Rao, M.~Wulfmeier, R.~Hadsell, and I.~Posner, ``Attention
  privileged reinforcement learning for domain transfer,'' 2020. [Online].
  Available: \url{https://openreview.net/forum?id=HygW26VYwS}
\BIBentrySTDinterwordspacing

\bibitem{andrychowicz2020learning}
O.~M. Andrychowicz, B.~Baker, M.~Chociej, R.~Jozefowicz, B.~McGrew,
  J.~Pachocki, A.~Petron, M.~Plappert, G.~Powell, A.~Ray, \emph{et~al.},
  ``Learning dexterous in-hand manipulation,'' \emph{The International Journal
  of Robotics Research}, vol.~39, no.~1, pp. 3--20, 2020.

\bibitem{james2017transferring}
S.~James, A.~J. Davison, and E.~Johns, ``Transferring end-to-end visuomotor
  control from simulation to real world for a multi-stage task,'' in
  \emph{Conference on Robot Learning}.\hskip 1em plus 0.5em minus 0.4em\relax
  PMLR, 2017, pp. 334--343.

\bibitem{liu2022distilling}
I.-C.~A. Liu, S.~Uppal, G.~S. Sukhatme, J.~J. Lim, P.~Englert, and Y.~Lee,
  ``Distilling motion planner augmented policies into visual control policies
  for robot manipulation,'' in \emph{Conference on Robot Learning}.\hskip 1em
  plus 0.5em minus 0.4em\relax PMLR, 2022, pp. 641--650.

\bibitem{rusu2016policy}
A.~A. Rusu, S.~G. Colmenarejo, C.~Gulcehre, G.~Desjardins, J.~Kirkpatrick,
  R.~Pascanu, V.~Mnih, K.~Kavukcuoglu, and R.~Hadsell, ``Policy distillation,''
  2016.

\bibitem{czarnecki2019distilling}
W.~M. Czarnecki, R.~Pascanu, S.~Osindero, S.~M. Jayakumar, G.~Swirszcz, and
  M.~Jaderberg, ``Distilling policy distillation,'' 2019.

\bibitem{chen2022system}
T.~Chen, J.~Xu, and P.~Agrawal, ``A system for general in-hand object
  re-orientation,'' in \emph{Conference on Robot Learning}.\hskip 1em plus
  0.5em minus 0.4em\relax PMLR, 2022, pp. 297--307.

\bibitem{kingma2013auto}
D.~P. Kingma and M.~Welling, ``Auto-encoding variational bayes,'' \emph{arXiv
  preprint arXiv:1312.6114}, 2013.

\bibitem{stone2021distracting}
A.~Stone, O.~Ramirez, K.~Konolige, and R.~Jonschkowski, ``The distracting
  control suite -- a challenging benchmark for reinforcement learning from
  pixels,'' 2021.

\bibitem{hansen2021generalization}
N.~Hansen and X.~Wang, ``Generalization in reinforcement learning by soft data
  augmentation,'' 2021.

\bibitem{agarwal2021deep}
R.~Agarwal, M.~Schwarzer, P.~S. Castro, A.~C. Courville, and M.~Bellemare,
  ``Deep reinforcement learning at the edge of the statistical precipice,''
  \emph{Advances in Neural Information Processing Systems}, vol.~34, 2021.

\bibitem{coleman2014reducing}
D.~Coleman, I.~Sucan, S.~Chitta, and N.~Correll, ``Reducing the barrier to
  entry of complex robotic software: a moveit! case study,'' 2014.

\bibitem{mittal2023orbit}
M.~Mittal, C.~Yu, Q.~Yu, J.~Liu, N.~Rudin, D.~Hoeller, J.~L. Yuan, R.~Singh,
  Y.~Guo, H.~Mazhar, A.~Mandlekar, B.~Babich, G.~State, M.~Hutter, and A.~Garg,
  ``Orbit: A unified simulation framework for interactive robot learning
  environments,'' \emph{IEEE Robotics and Automation Letters}, vol.~8, no.~6,
  pp. 3740--3747, 2023.

\bibitem{IsaacDeveloper}
\BIBentryALTinterwordspacing
NVIDIA, ``Nvidia isaac sim.'' [Online]. Available:
  \url{https://developer.nvidia.com/isaac-sim}
\BIBentrySTDinterwordspacing

\end{thebibliography}

\end{document}